\documentclass[conference]{IEEEtran}
\IEEEoverridecommandlockouts
\usepackage{cite}
\usepackage{amsmath,amssymb,amsfonts}
\usepackage{dsfont}
\usepackage{subfigure}
\usepackage{algorithm}
\usepackage{algpseudocode}
\usepackage{graphicx}
\usepackage{textcomp}
\usepackage{xcolor}

\DeclareMathOperator*{\argmax}{argmax}

\def\BibTeX{{\rm B\kern-.05em{\sc i\kern-.025em b}\kern-.08em
    T\kern-.1667em\lower.7ex\hbox{E}\kern-.125emX}}
\begin{document}

\title{Anomaly Detection for Scalable Task Grouping in Reinforcement Learning-based RAN Optimization\\
}

\author{\IEEEauthorblockN{\IEEEauthorrefmark{2}Jimmy Li\IEEEauthorrefmark{1}, \IEEEauthorrefmark{2}Igor Kozlov\IEEEauthorrefmark{1}, Di Wu\IEEEauthorrefmark{1}, Xue Liu\IEEEauthorrefmark{1}, Gregory Dudek\IEEEauthorrefmark{1}}
\IEEEauthorblockA{\IEEEauthorrefmark{1}\{jimmy.li,i.kozlov,di.wu1,steve.liu,greg.dudek\}@samsung.com, \IEEEauthorrefmark{2}equal contribution}%
\thanks{This work was done at the Samsung AI Center - Montreal. 1000 Sherbrooke West, Montreal, Quebec, Canada.}
}

\maketitle

\begin{abstract}
The use of learning-based methods for optimizing cellular radio access networks (RAN) has received increasing attention in recent years. This coincides with a rapid increase in the number of cell sites worldwide, driven largely by dramatic growth in cellular network traffic. Training and maintaining learned models that work well across a large number of cell sites has thus become a pertinent problem. This paper proposes a scalable framework for constructing a reinforcement learning policy bank that can perform RAN optimization across a large number of cell sites with varying traffic patterns. Central to our framework is a novel application of anomaly detection techniques to assess the compatibility between sites (tasks) and the policy bank. This allows our framework to intelligently identify when a policy can be reused for a task, and when a new policy needs to be trained and added to the policy bank. Our results show that our approach to compatibility assessment leads to an efficient use of computational resources, by allowing us to construct a performant policy bank without exhaustively training on all tasks, which makes it applicable under real-world constraints.
\end{abstract}


\section{Introduction}
Reinforcement learning (RL) has been shown to be an effective approach for optimizing cellular radio access networks (RAN) \cite{mwanje2016cognitive,li2022traffic}. 
However, most existing methods use RL agents to control a single site or a small set of neighboring sites. Deployment of RL controllers across large geographic regions involving hundreds or thousands of sites poses key challenges for network operators. Due to the variations in traffic patterns near different sites, it is challenging to train a single RL control policy that works well across all sites. Different sites can be seen as different tasks, and it is well-known that multitask learning can lead to degraded performance when dissimilar or conflicting tasks compete for model capacity \cite{fifty2021efficiently}. On the other hand, training and maintaining separate RL policies for every site is expensive: RL policies typically require a large amount of environment interactions to train. Furthermore, treating each site independently does not allow policies of similar sites to benefit from positive knowledge transfer.

In our prior work, we have proposed a method for creating an RL policy bank that can perform load balancing in RANs across a large variety of traffic patterns \cite{li2022traffic}. The policy bank contains a small set of policies, each of which caters to multiple similar sites. Our method first trains RL policies independently for all sites, and then iteratively merges together similar policies to yield a concise policy bank. 
While this approach enables positive knowledge transfer between similar sites and reduces the maintenance overhead by keeping the number of policies small, it requires training on all sites, which can be prohibitively expensive if the number of sites is large.

In this paper, we propose a scalable training framework that constructs an RL policy bank for a large number of tasks. The specific problem we consider is inter-frequency load balancing in cellular networks, and in this context, the set of tasks refers to the set of cell sites with varying local traffic patterns. However, our framework can generally be used under any multitask RL setting, and can be applied to other RAN optimization problems. 
Our framework processes the set of tasks sequentially, and continuously updates the policy bank. 
Central to our framework is a mechanism for assessing the compatibility between the task being processed and the existing policies.
We only train new RL policies for tasks that are incompatible with all existing policies, and we forgo training on tasks that our policy bank can already handle, thereby leading to significant computational savings. This overall process can be viewed as grouping together similar tasks that can be handled by the same policy.
To prevent unbounded growth in the number of policies, we use knowledge distillation to periodically merge the most similar policies such that the total bank size remains capped.
During deployment, a policy selector can be used to automatically choose the most suitable policy for the target task based on current observations, as we have shown in prior work \cite{xu2023policy}.

Determining whether a task is compatible with an existing policy is a key problem addressed by this paper. Due to variations in task difficulty, the maximum attainable reward varies between tasks, which makes it challenging to determine compatibility based solely on reward.
We propose to formulate compatibility assessment as an anomaly detection (AD) problem. AD typically aims to identify whether samples originate from the same distribution. 
In the context of our framework, we aim to determine whether the interaction experiences (i.e. time series of observed states the system evolved through) obtained during RL policy training belong to the same distribution as those obtained when evaluating the trained policy on new tasks. 
Due to the time series nature of the samples, this type of AD is related to behavior change detection, or more commonly referred to as change point detection (CPD), which we elaborate in sections \ref{sec:ad_in_rans} and \ref{sec:compat_ad}. 
Our general framework allows any existing AD technique to be used to assess compatibility, as long as it is able to output a similarity score between two time series. 
We demonstrate that our AD-based approach leads to significant computational savings when compared to conventional rule-based approaches of determining compatibility based on key performance indicators (KPIs) of the network, achieving equal or better performance while requiring substantially less computation cost.

\section{Background}

\subsection{Load balancing in cellular networks} \label{sec:lb}
\label{sec:background-load-balancing}
This paper focuses on the load balancing problem in cellular networks. We consider a telecommunication network where each site or base station hosts a number of \textbf{cells}, each of which serves user equipment (UE) at a specific carrier frequency. The cells at a site are grouped into non-overlapping direction ranges, known as \textbf{sectors}. Our work focuses on inter-frequency load balancing, which moves UEs between cells in a sector.

We consider two load balancing mechanisms. The first is active UE load balancing, which uses handovers to move transmitting UEs between cells. The handover condition is $\mbox{RSRP}_j > \mbox{RSRP}_i + \alpha_{i,j} + H$,
where $\mbox{RSRP}_i$ is the Reference Signal Received Power (RSRP) of the current serving cell, $\mbox{RSRP}_j$ is the RSRP of a target cell $\mbox{RSRP}_j$, $\alpha_{i,j}$ is the triggering threshold for handover from cell $i$ to $j$, and $H$ is the hysteresis. 

The second load balancing mechanism is idle UE load balancing. Cell reselection is used to change the camped cell of idle UEs, which reduces handover delays when the UE becomes active. The condition for cell reselection is
$\mbox{RSRP}_i < \beta_{i,j} \mbox{ and } \mbox{RSRP}_j > \lambda_{i,j}$,
where $\beta_{i,j}$ and $\lambda_{i,j}$ are the thresholds for changing the camped cell from $i$ to $j$. Load balancing is achieved by optimizing the parameters $\alpha_{i,j}$, $\beta_{i,j}$ and $\lambda_{i,j}$ for all cell pairs.

\subsection{Reinforcement learning}
RL is a machine learning paradigm, in which an agent aims to learn a control policy $\pi$ via interacting with the environment to maximize its long-term expected reward.
Recently, there has been growing interest in applying RL to cellular network optimization \cite{wu2023reinforcement}. Model-free RL methods such as Q-learning have been a popular approach, since it allows us to avoid modelling the system dynamics of the environment, and rely on the trained policy to directly output control actions based on observations
\cite{mwanje2016cognitive}.
Model-free deep RL, which uses a deep network to model the policy, has received increasing attention lately 
\cite{wu2023reinforcement}. 
Unlike Q-learning, deep RL can support continuous state and action spaces, allowing it to address a larger range of problems.
Our prior work has shown that model-free deep RL methods can achieve substantial performance improvements over hand-engineered methods \cite{li2022traffic}, and is the approach we use in this paper as well.

\subsection{Anomaly detection in cellular networks}
\label{sec:ad_in_rans}

AD has been used extensively in cellular networks to detect anomalous events, such as service interruptions, performance degradations, and security threats \cite{ad-cell-survey}. 
A common approach involves detecting irregularities in an expert chosen subset of the system's state space, called key performance indicators (KPIs) 
\cite{barreto20053G}.

A variety of methods have been applied to AD, including clustering algorithms \cite{barreto20053G}, 
correlation analysis \cite{munoz2016}, 
auto-encoders \cite{chawla2020interpretable}. Change point detection (CPD) \cite{nonmlcpdsreview} is commonly used for AD when data samples represent extended sequences of observations over time. In CPD, one calculates the distance between two time series samples in order to determine whether they belong to the same distribution. This is highly related to our problem of compatibility assessment, which compares two time series consisting of RL interaction experiences.
One classical robust and high performing approach is binary segmentation \cite{binseg}, which sequentially splits a time series at points such that the difference between the backwards and forward-looking segments is maximized. Another common method is to quantify the mismatch between consecutive non-overlapping windows of a fixed length.  This approach is often enhanced by applying representation learning techniques - a field of Machine Learning (ML) that extracts patterns such as those related to periodicity.  This is especially suitable to seasonal data, like in RANs. In our own prior work, we have developed a state-of-the-art method, TREX-DINO (Time series REpresentation eXtraction using DIstillation with NO labels), which is trained in a self-supervised manner to capture patterns that are hard to describe in human language \cite{kozlov2023self}. 

\section{Method}

This section describes our scalable approach to training an RL policy bank for a large number of tasks.
We refer to our method as a task grouping framework, since the underlying process groups together similar tasks and trains a shared policy for each group.
We first present the overall framework structure, followed by in-depth discussions of the key components.

\subsection{Scalable task grouping framework}\label{sec:framework}
Our framework is shown in
Algorithm \ref{alg:framework}. The input consists of a list of tasks $D$, the maximum policy bank size $n$, and a sample size $k$ that controls the number of tasks to sample per iteration. Setting $k > 1$ allows the implementation to processes multiple tasks in parallel.

During initialization (lines 1-6), we sample $k$ tasks from $D$, and train an RL policy for each task to bootstrap the policy bank $\Pi$. The RL  formulation is further discussed in Section \ref{sec:rl}. We evaluate all trained policies on the training task to generate ``training interaction experiences" that can be used later for compatibility assessment. Our implementation uses the sequence of observed states as interaction experience, but more generally, the action and reward can also be used.

After initialization, we iteratively sample $k$ tasks at a time (line 8-9), until all tasks are processed.
In each iteration, we first cross-evaluate all policies in the policy bank on all the sampled tasks (line 10) to generate ``test interaction experiences". For each policy and sampled task, we invoke an AD-based method to generate a compatibility score by comparing the policy's training interaction experience and the test interaction experience for the sampled task (line 11). A threshold is applied to determine whether the sampled task is compatible with the policy. AD-based compatibility assessment is further discussed in Section \ref{sec:compat_ad}. 

Tasks with at least one compatible policy are ``grouped" with the most compatible policy based on the compatibility scores (line 13). New policies are trained for tasks that are incompatible with all existing policies (lines 15). If the total number of policies is larger than $n$, we use knowledge distillation to repeatedly merge the most similar policies until the number of policies is $n$ (lines 18-22). Policy distillation is further discussed in Section \ref{sec:distill}.

\algnewcommand\algorithmicforeach{\textbf{for each}}
\algdef{S}[FOR]{ForEach}[1]{\algorithmicforeach\ #1\ \algorithmicdo}
\algrenewcommand\algorithmicrequire{\textbf{Input:}}
\algrenewcommand\algorithmicensure{\textbf{Output:}}
  
\begin{algorithm}
\caption{Scalable task grouping}\label{alg:framework}
\begin{algorithmic}[1]
\Require{ tasks $D$, max policy bank size $n$, sample size $k$ }
\Ensure{ Policy bank $\Pi$ }
\State $\Pi \gets$ empty policy bank
\State $d \gets$ random sample of $k$ tasks from $D$
\State $D \gets D \setminus d$
\State Train policies $\pi_1,...\pi_k$ for each task in $d$
\State Evaluate each $\pi_i$ on training task, save interaction experience
\State $\Pi \gets \Pi \cup \{\pi_1,...\pi_k\}$
\While{$D$ is not empty}
    \State $d \gets$ random sample of $k$ tasks from $D$
    \State $D \gets D \setminus d$
    \State Evaluate each $\pi \in \Pi$ on each task in $d$, save interaction \hspace*{8mm}experience
    \State Generate compatibility score between each $\pi \in \Pi$ and \hspace*{8mm}each task in $d$
    \State $c \gets$ tasks in $d$ compatible with at least one $\pi \in \Pi$
    \State Group each task in $c$ with the most compatible $\pi \in \Pi$ \hspace*{8mm}based on compatibility score
    \State $u \gets$ tasks in $d$ incompatible with all $\pi \in \Pi$
    \State Train policies $\pi_1,...\pi_{|u|}$ for each task in $u$
    \State Evaluate each newly-trained $\pi_i$ on training task, \hspace*{8mm}save interaction experience
    \State $\Pi \gets \Pi \cup \{\pi_1,...\pi_{|u|}\}$
    \While{$|\Pi| > n$}
        \State Identify most similar policies $(\pi_i, \pi_j) \in \Pi$
        \State Distill $(\pi_i, \pi_j)$ into single policy $\pi^*$
        \State $\Pi \gets \Pi \setminus \{\pi_i, \pi_j\} \cup \pi^*$
    \EndWhile
\EndWhile
\end{algorithmic}
\end{algorithm}

\subsection{RL policy training}\label{sec:rl}
Our proposed framework constructs a bank of RL policies. Applying RL requires the formulation of a Markov decision process (MDP) that specifies the state and action spaces of the RL agent, as well as the reward signal that guides learning. 
Generally, any MDP formulation can be used in conjunction with our framework. 
Here, we present our MDP formulation for load balancing.

Each state $s_t \in \mathcal{S}$ at time step $t$ is a vector containing the number of active UEs in each cell, the bandwidth utilization of each cell, and the average throughput of each cell. These values are averaged over the duration of each time step.
Each action $a_t \in \mathcal{A}$ taken at time $t$ consists of the load balancing parameters $\alpha_{i,j}, \beta_{i,j}$ and $\lambda_{i,j}$ as discussed in Section~\ref{sec:lb}. 

The reward function $\mathcal{R}$ outputs a weighted average of several key performance indicators (KPIs) that are computed based on throughput, which is a part of the state. Letting $x_i$ be the throughput of the i-th cell, and $N_c$ be the total number of cells, the KPIs are defined as follows:
\begin{itemize}
    \item $G_{\mbox{min}} = \min_{i\in \{1,..,N_c\}} x_i $ is the minimum throughput among all cells. Higher is better.
    \item $G_{\mbox{avg}} = \frac{1}{N_c} \sum_{i=1}^{N_c} x_i$ is the average throughput over all cells. Higher is better.
    \item $G_{\mbox{sd}} = \sqrt{\frac{1}{N_c} \sum_{i=1}^{N_c}\left(x_i-G_{\mbox{avg}}\right)^2}$ is the standard deviation in throughput among all cells. Lower is better, since we aim to evenly distribute load among the cells.
    \item $G_{<\chi} = \sum_{i=1}^{N_c} \mathds{1}(x_i < \chi)$ is the number of cells with throughput lower than a threshold $\chi$. Lower is better.
\end{itemize}
Finally, the reward is calculated as $r_t = \phi_1 G_{avg} + \phi_2 G_{min} + \phi_3 (1+G_{sd})^{-1} + \phi_4 (N_c - G_{\chi})$, where the coefficients $\phi_i$ are hyperparameters chosen as in \cite{li2022traffic}. 
Standard RL algorithms can be used to train a policy $\pi_i(a_t|s_t)$ that aims to maximize the expected future rewards.
We used (PPO)~\cite{schulman2017ppo}.

\subsection{Compatibility assessment via anomaly detection}
\label{sec:compat_ad}

Given a policy's training interaction experience $\tau_1$ and a test interaction experience $\tau_2$ generated by the same policy on a new task, we apply AD, or more specifically CPD methods, to measure the difference between  $\tau_1$ and $\tau_2$, which we then use to quantify the compatibility between the policy and the new task. In our implementation, each $\tau$ is a sequence of state vectors. We assume for now that each state vector is single dimensional, and will address the multi-dimensional case at the end of this section. 

We implement two state-of-the-art AD techniques for performing CPD in RANs: a non-ML and a ML-based approach.
The non-ML AD technique is based on binary segmentation (BG) \cite{binseg}. We concatenate $\tau_1$ and $\tau_2$, and apply BG on the combined series $\bar{\tau}$. BG defines the gain at time $t$: $G(t) \equiv C(\bar{\tau}_{0..T}) - C(\bar{\tau}_{0..t}) - C(\bar{\tau}_{t..T})$,
where $T$ is the length of $\bar{\tau}$, $C$ is the cost function, which in our implementation is computed using the radio basis function kernel as in \cite{ruptures}. We then compute $t^*=\argmax_t G(t)$ as the change point. 
The series is then split at $t^*$ and we repeat the process on the sub-series. We continue this process to obtain gain for all $t \in \bar \tau$, and then perform peak detection as in \cite{de2021change} to find the most likely candidate for the change point. The obtained likelihood is used as the compatibility score. 

The ML-based technique we implement is TREX-DINO (TD), which we have introduced in our prior work \cite{kozlov2023self}.
This approach is based on a self-supervised learning technique originally proposed for representation learning on images \cite{dino}. 
For the image domain, this approach trains two separate self-attention networks by presenting them with differently augmented views of the same image \textemdash a technique known as ``multi-crop''. 
One network (teacher) receives only big scale crops covering more than 50\% of the image, while the other network (student) receives crops of all sizes. 
The student $(s)$ is trained to mimic the output of the teacher $(t)$ by minimizing the aggregation of cross-entropy ($H$) terms calculated over all pairwise multi-crop combinations ($V$) sampled without replacement: 
$\sum_{x_t, x_s \in V, x_s \neq x_t} H(P_t(x_t), P_s(x_s))$,
where $x$ is the input and $P$ is the output of the corresponding network. 
The teacher's network weights are updated using an exponential moving average of the student's weights in a self-distillation manner, which makes representations of both networks similar after substantive training. 
In TREX-DINO (TD), this approach is extended to the time-series domain, and is applied to CPD in RANs. 
In this context, multi-crops consist of segments of time series data. Additionally, Gaussian blur and additive Gaussian noise are used to augment time-series observations to improve model generalization. 
Notably, TD offers superior performance and robustness with respect to the accurate choice of threshold compared to the other CPD methods \cite{kozlov2023self}.

We pre-train TD prior to the task grouping process on a dataset consisting of RAN states generated using default fixed configuration parameters \cite{kozlov2023self}. 
Importantly, this data is different from the time series obtained during task grouping, as those generated by trained RL policies, which actively change the configuration parameters. 
As in the vision domain \cite{dino}, TD learns emerging properties in historical observations in RAN data and can efficiently measure similarity between the time series evolution of a system using the cosine distance in the embedding space $d(\tau_1, \tau_2) = 1 - \langle P(\tau_1) \cdot P(\tau_2) \rangle$, where $\langle . \rangle$ is a dot product. 
We note that due to self-distillation, the outputs of the student and teacher networks nearly coincide. 

To handle multi-dimensional states, for both BG and TD, our implementation first computes the compatibility score separately for each dimension, giving us a vector $v$ of scores. Then, the L2 norm of $v$ is the final compatibility score. 
Additionally, we apply standard normalization procedures to all time series $\bar \tau$, in which we take the logarithimic transformation, and then normalize the data to have zero mean and unit variance.

\subsection{Policy merging via distillation} \label{sec:distill}
Whenever the number of policies in our policy bank exceeds the maximum desired size $n$, our framework uses knowledge distillation 
to iteratively merge together the most similar RL policies until the maximum size is respected. 
Our method is based on a policy distillation technique presented in our own prior work \cite{li2022traffic}. 
We define the similarity between two policies $\pi_i$ and $\pi_j$ as
$\delta(i,j) = \mathbb{E}_{s \in S_j} ||\argmax_a \pi_i(a|s) - \argmax_a \pi_j(a|s)||_2$,
where the state vectors $S_j$ comes from the saved interactions when evaluating $\pi_j$ on the task that it is trained on. Intuitively, $\delta(i,j)$ measures the average L2 distance between the action produced by the two policies given the same input state.
To merge together two policies $\pi_i$ and $\pi_j$, we treat them as teachers, and train a student policy $\pi^*$ that mimics both teachers. The training loss is
$J = \sum_{s \in S_i}D_{kl}\big(\pi_i(a|s)||\pi^*(a|s)\big) + \sum_{s \in S_j}D_{kl}\big(\pi_j(a|s)||\pi^*(a|s)\big)$,
where $D_{kl}$ denotes Kullback–Leibler (KL) divergence.
The student policy also inherits both parents' training interaction experiences, which are used for compatibility assessment, or if the student policy is merged again with another policy. Please refer to our prior work \cite{li2022traffic} for more details.

\begin{figure}[htbp]
\centering
\subfigure[Performance, varying compatibility thresholds]
{\includegraphics[width=.85\columnwidth]{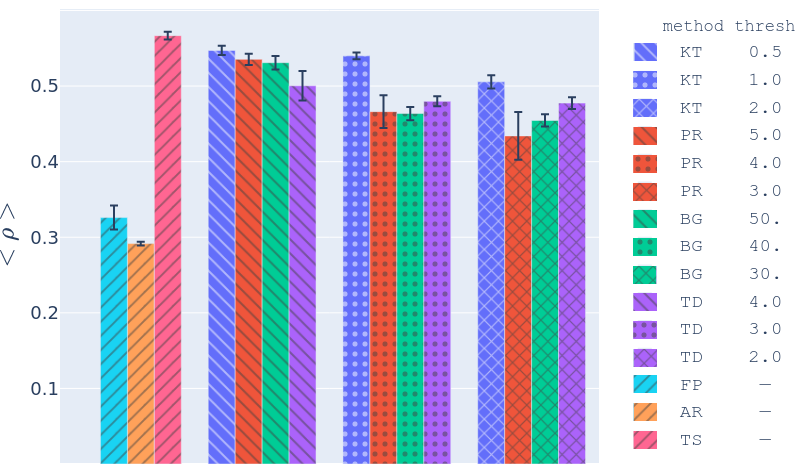}}
\subfigure[Performance to training ratio, varying compatibility thresholds]
{\includegraphics[width=.85\columnwidth]{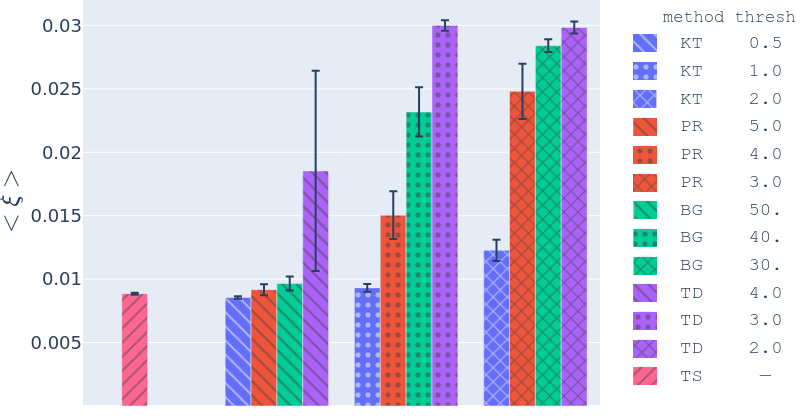}}
\subfigure[Performance, varying bank sizes]
{\includegraphics[width=.85\columnwidth]{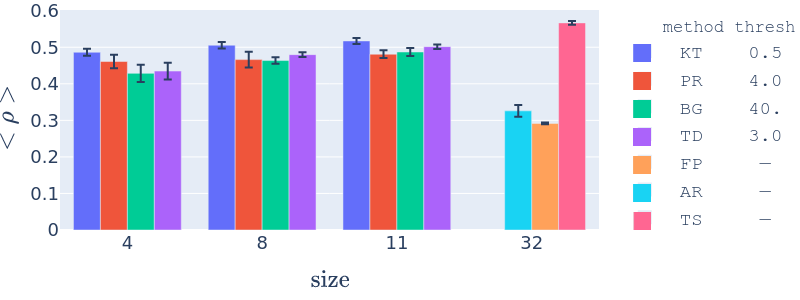}}
\subfigure[Performance to training ratio, varying bank sizes]
{\includegraphics[width=.85\columnwidth]{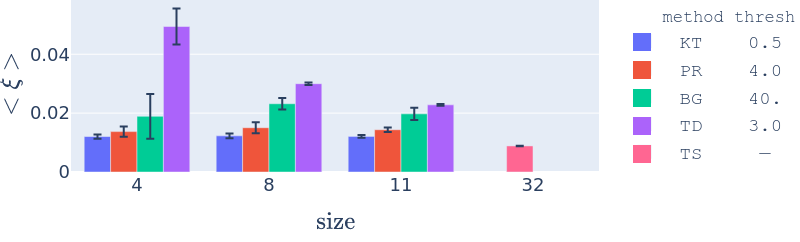}}
\caption{Performance $\rho$ and performance to training ratio $\xi$ for all methods across different hyperparameter settings. Error bars denote standard deviation across 5 different random seeds. Our AD-based approach (BG, TD) is capable of achieving much higher $\xi$ while closely matching the performance of the non-AD-based KT, PR, and TS, which demonstrates the superior efficiency of our method.
}
\label{fig:ratios}
\end{figure}

\section{Experimental Results}
\subsection{Simulation environment}\label{sec:experiment_setup}

We evaluate our framework using a proprietary system level simulator for 4G/5G cellular networks as in our prior work \cite{li2022traffic,kozlov2023self}. Each simulation instantiates 7 sites: a center site surrounded by 6 additional sites. Each site has 3 sectors, and each sector has 4 cells. The RL agent is expected perform inter-cell load balancing for one sector in the center site. Additional sites introduce interference. 
We use 32 sets of simulation parameters to simulate different traffic scenarios that constitute different tasks. These parameters are fit to reproduce traffic data observed at 32 real-world cell sites. 

\subsection{Evaluation metric}

To compute the performance $\rho$ for a given policy bank $\Pi$, we evaluate each $\pi \in \Pi$ on all tasks grouped with it (see Section \ref{sec:framework}), as well as the task used to train $\pi$.
Then, $\rho$ is the mean cumulative reward across all of these tasks. Let $w$ be the total environment interaction steps across all training sessions when constructing $\Pi$. We define  performance to training ratio as $ \xi = \frac{\rho}{w}$.
In our implementation, each policy is trained with 200,000 interaction steps. 
$\xi$ describes performance gain per 100,000 interaction steps, which quantifies the trade-off between performance and computation cost.

\subsection{Baselines and ablation parameters}

We compare our proposed approach with several non-AD baselines:
\begin{itemize}
    \item FP: Fixed set of control parameters used for all tasks \cite{li2022traffic}.
    \item AR: Adaptive rule-based method that selects load balancing parameters based on cell load \cite{yang2012high}.
    \item TS: Train separate task-specific policies for every task, which represents the highest computation cost. This can also be considered as an upper bound on performance, since the policies do not need to generalize across multiple tasks.
    \item KT: Rule-based variant of our framework, where we consider a task to be compatible with policy $\pi$ if the KPI $G_{\mbox{min}}$ (see Section \ref{sec:rl}) surpasses a threshold for all time steps when executing $\pi$ on the task. The lowest $G_{\mbox{min}}$ among all time steps is the compatibility score.
    \item PR: the Pearson correlation coefficient, $r$, is used to define the compatibility score between two time-series as $\sqrt{1 - r^2}$.
\end{itemize}
We consider the two AD-based variants of our framework:
\begin{itemize}
\item BG: Binary segmentation for compatibility assessment.
\item TD: TREX-DINO for compatibility assessment.
\end{itemize}
For the methods presented above, we used default settings and ablate our studies over two key hyperparameters: policy bank size $n$ and compatibility threshold.
The bank size affects learning capacity: smaller banks requires policies to achieve greater generalization. 
We conduct runs with $n=4,8,11,32$. 
We set the sample size $k$ to be equal to $n$ for all runs.

The compatibility threshold, when tight, considers all tasks incompatible, which leads to excessive training. 
A loose threshold uses one model for all tasks, which negatively affects performance. 
KT's threshold is chosen using expert knowledge and is varied by a factor of two to cover a substantial part of the hyperparameter space. 
When choosing thresholds for PR, BG, and TD, we sample all aforementioned policy bank sizes, and use the median distance of pairwise task compatibility scores as the default value. 
Notably, this result matches the heuristic value for BG \cite{ruptures}. 
Two additional methods' thresholds (extended from the default one by one standard deviation) are introduced for assessing robustness. 
All thresholds are rounded to one significant figure. 

All methods are evaluated across 5 different random seeds that introduce randomness to the simulated UE behavior.
For methods based on our task grouping framework, the random seed also affects the order in which tasks are sampled.

\begin{figure}[t]
\centering
\subfigure[Number of trained tasks]
{\includegraphics[width=.8\columnwidth]{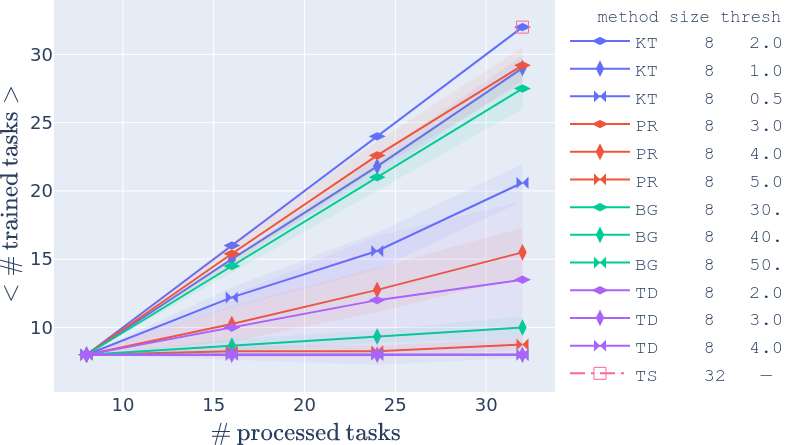}}
\subfigure[Performance on processed tasks]
{\includegraphics[width=.8\columnwidth]{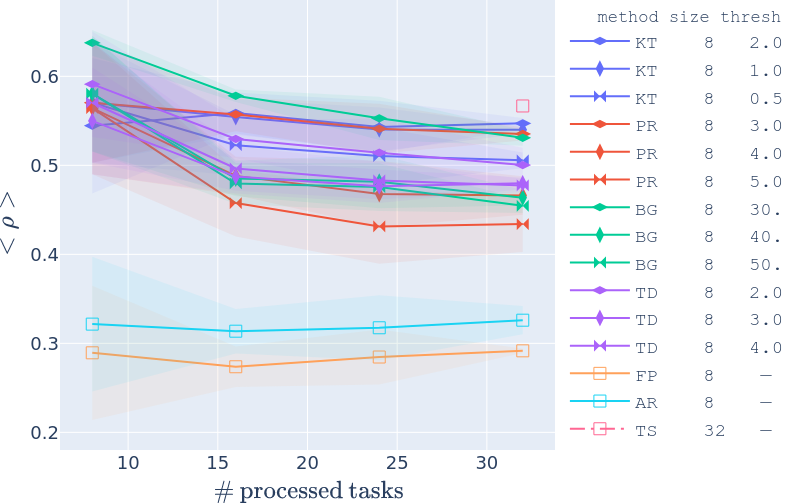}}
\caption{Number of trained tasks and performance $\rho$ as a function of total tasks processed so far, across different compatibility thresholds. Ribbons denote standard deviation across 5 different random seeds. Our AD-based methods (BG, TD) can closely match the performance of KT and PR with less training.}
\vspace{-.5cm}
\label{fig:performance_train_time_thresholds}
\end{figure}

\begin{figure}[htbp]
\centering
\subfigure[Number of trained tasks]
{\includegraphics[width=.8\columnwidth]{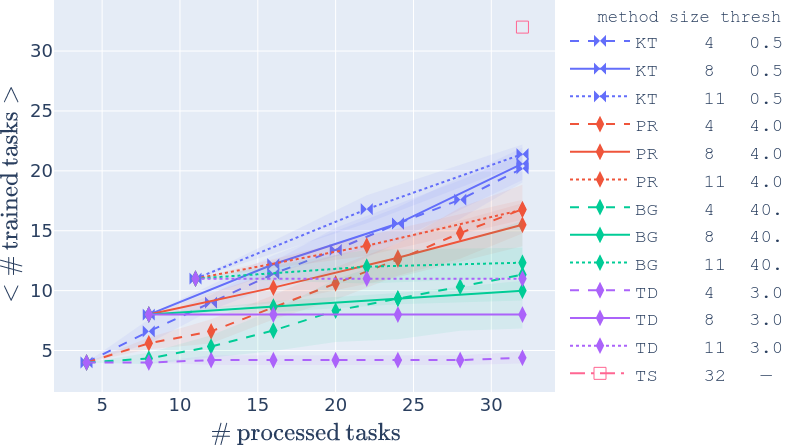}}
\subfigure[Performance on processed tasks]
{\includegraphics[width=.8\columnwidth]{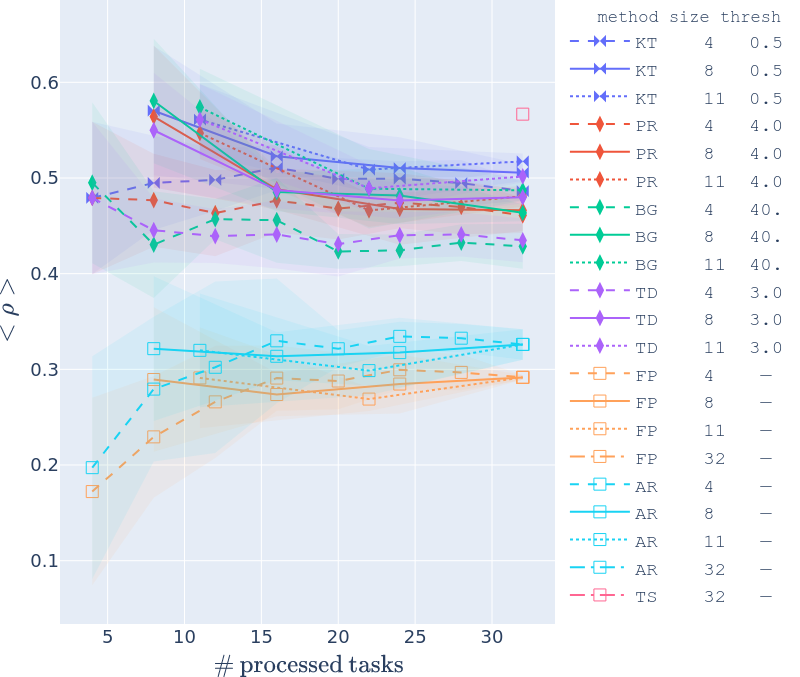}}
\caption{Same format as Figure \ref{fig:performance_train_time_thresholds}, except we vary the policy bank size instead of the compatibility threshold. Performance $\rho$ generally decreases as more tasks are processed, and as bank size decreases, since this requires each policy to generalize to more tasks.}
\vspace{-.5cm}
\label{fig:performance_train_time_bank_size}
\end{figure}

\subsection{Results}
Figures \ref{fig:ratios} (a) and (b) show the performance $\rho$  and the performance to training ratio $\xi$ of all methods. FP and AR have the lowest performance, and since they do not involve any training, the ratio is not applicable to them. The task-specific policies have the highest performance as expected, but a low ratio indicating poor computational efficiency.

Regarding the methods based on our framework, our proposed AD-based method (BG, TD) generally achieves significantly higher performance to training ratios, while closely matching (KT, PR) in performance. 
This suggests that AD allows to group together similar tasks better than conventional rule-based and correlation approaches. 

The ablation studies in Figures \ref{fig:ratios} (c) and (d) suggest that while threshold choice impacts performance, it is evident that AD methods, and especially TD, provides higher robustness and significantly better computational efficiency with minimal sacrifices in performance.
In addition, methods similar to KT and PR are more sensitive to outliers typical in industrial settings: an occasional dip in performance can trigger training, even when it is not necessary.

Figures \ref{fig:performance_train_time_thresholds} and \ref{fig:performance_train_time_bank_size} provide additional insight into the task grouping process of our framework. We plot the cumulative amount of RL training and average performance across processed tasks at each iteration of our method. 
As reference, we also plot the performance of conventional baselines (FP, AR) by evaluating them on the same set of tasks that have already been processed by our framework. 
Overall, our earlier conclusions
are also applicable to the intermediate stages of the policy bank construction: our AD based method (BG, TD) generally invokes training more sparingly than the non-AD ones (KT, PR) as tasks are processed, but can achieve comparable performance with less training. We also see that the policy bank's performance generally decreases as more tasks are processed, since this requires the policies to generalize across more tasks, which is expected.

\section{Conclusion}
We have presented a framework for training an RL policy bank across a large number of tasks. By  using AD to identify tasks that are incompatible with existing policies, our framework only trains on incompatible tasks, which leads to significant computational savings while maintaining good performance. We have applied our framework to the load balancing problem in cellular networks, and have demonstrated its effectiveness and applicability to real-world systems.
Our framework generalizes to other RAN optimization tasks (i.e. energy saving), and other multi-task RL problems.

\bibliographystyle{IEEEtran}
\bibliography{main.bib}

\end{document}